\def\year{2021}\relax
\documentclass[letterpaper]{article} 
\usepackage{aaai21}  
\usepackage{times}  
\usepackage{helvet} 
\usepackage{courier}  
\usepackage[hyphens]{url}  
\usepackage{graphicx} 

\usepackage{bm} 
\usepackage{booktabs} 

\usepackage{natbib}  
\usepackage{caption} 
\usepackage{subfigure}
\usepackage{amsmath}
\usepackage{multirow}
\usepackage{url}

\title{
C-Watcher: A Framework for Early Detection of High-Risk Neighborhoods\\ Ahead of COVID-19 Outbreak
}

\author{
Congxi Xiao$^{1,2*}$, Jingbo Zhou$^{2\dagger*}$, Jizhou Huang$^{2}$\thanks{C. Xiao, J. Zhou and J. Huang contributed equally to the paper. This work was done when C. Xiao was an intern at Baidu Inc.}, An Zhuo$^{2}$, Ji Liu$^{2}$, Haoyi Xiong$^{2}$,
Dejing Dou$^{2}$\thanks{J. Zhou and D. Dou are  the corresponding authors.}}
\affiliations{
$^{1}$University of Science and Technology of China,
$^{2}$Baidu Inc., China\\
\{v\_xiaocongxi, zhoujingbo, huangjizhou01, zhuoan, liuji04, xionghaoyi, doudejing\}@baidu.com
}

\begin{document}
\maketitle
\begin{abstract}
The novel coronavirus disease (COVID-19) has crushed daily routines and is still rampaging through the world. 
Existing solution for nonpharmaceutical interventions usually needs to timely and precisely select a subset of residential urban areas for containment or even quarantine, where the spatial distribution of confirmed cases has been considered as a key criterion for the subset selection. 
While such containment measure has successfully stopped or slowed down the spread of COVID-19 in some countries, it is criticized for being inefficient or ineffective, as the statistics of confirmed cases are usually time-delayed and coarse-grained.
To tackle the issues, we propose C-Watcher, a novel data-driven framework that aims at screening every neighborhood in a target city and predicting infection risks, prior to the spread of COVID-19 from epicenters to the city. In terms of design, C-Watcher collects large-scale long-term human mobility data from Baidu Maps, then characterizes every residential neighborhood in the city using a set of features based on urban mobility patterns.
Furthermore, to transfer the firsthand knowledge (witted in epicenters) to the target city before local outbreaks, we adopt a novel adversarial encoder framework to learn ``\emph{city-invariant}'' representations from the mobility-related features for precise early detection of high-risk neighborhoods, even before any confirmed cases known, in the target city. We carried out extensive experiments on C-Watcher using the real-data records in the early stage of COVID-19 outbreaks, where the results demonstrate the efficiency and effectiveness of C-Watcher for early detection of high-risk neighborhoods from a large number of cities. 
\end{abstract}

\section{Introduction}
The novel coronavirus disease (COVID-19), which has been officially announced as a pandemic by the World Health Organization (WHO), is perhaps the most serious public health emergency over the past decades. The coronavirus continues to spread around the globe, which challenges the governments and medical systems all over the world.

While metropolitan-wide lockdown has demonstrated its effectiveness as a nonpharmaceutical intervention in several countries, the cost of such measures, including unemployment, economic crash, and social anxiety, makes it a tough decision on behalf of administrators. A compromised solution is to place containment measures onto a subset of areas in a city to stop or slow down the spread of COVID-19 while minimizing the social and economic cost. To precisely distinguish the high-risk areas from the city, the spatial distribution of confirmed cases has been used as the key criterion for the potential containment measures in a data-driven fashion.

While the spatial statistics of confirmed cases work, the time consumption and the granularity of data acquisition significantly lower the efficiency and effectiveness of such methods.
For example, the incubation period of COVID-19 is around 5--6 days on average, but it could last as long as 14 days. During such a period, a community or neighborhood would have been already invaded by a small number of asymptotic carriers who might not be with any clinical symptoms.  
When a small number of cases being confirmed, the community and its surrounding neighborhoods may have fallen into COVID-19 for a long time. Furthermore, though the mobility of confirmed patients is usually well restricted, the asymptomatic carriers with no symptoms would still spread the virus to where he/she has gone. When fine-grained mobility traces are not available, the administrators can only place containment measures to places in coarse-grained.

To tackle the technical issues, in this paper, we propose C(OVID)-Watcher, a novel framework to support early detection of high-risk residential neighborhoods for fighting against the spread of COVID-19. Our intuition is to incorporate human mobility data, so as to (1) characterize the socioeconomic and demographic status of every neighborhood~\cite{borjas2020demographic,huang2020kddtransp} based on ``how residents move''~\cite{renso2013you} and (2) unfold the spatial interactions~\cite{jiang2020spatial} and potential influences on COVID-19 caused by the mobility of massive asymptotic carriers. Specifically, C-Watcher includes a mobility data-driven machine learning model that screens every neighborhood in a target city and predicts the infection risks, prior to the spread of COVID-19 from epicenters to the city.

In addition to the use of mobility-related features, we also hope to generalize the evidence already witted in the epicenter for screening the risk of neighborhoods in the target city, prior to or in the early stage of local outbreaks. To achieve the goal, a core component of C-Watcher is a novel cross-city transfer learning model that transfers knowledge about COVID-19 infections from the epicenter to the target city, while cities are quite different in a large number of domains, ranging from living, foods, transportation, and residences.

All in all, we have made three contributions as follows.  
\begin{itemize}
    \item Through extensive data analytics, we explore a set of empirical features related to long-term/regular human mobility patterns (before the COVID-19). With such long-term mobility features, the socioeconomic and demographic status, as well as the spatial interactions among neighborhoods could be well characterized. In this way, one can easily distinguish high-risk neighborhoods from the urban area and predict the potential risks for infection, with respect to the two factors.
    
    \item While these features are with certain discriminative information for risk prediction, they also involve some city-specific characteristics. For example, the popular choices for transport modes in different cities vary. Such city-specific characteristics burden the use of mobility-related features to transfer the knowledge obtained in the epicenter to the target city. To generalize the knowledge transfer, C-Watcher adopts a novel adversarial encoder-decoder framework to learn the ``city-invariant'' representations from the mobility-related features for prediction.
    
    \item To validate C-Watcher, we collect and construct real-world datasets for high-risk neighborhood detection based on the publicly available information from the web and human mobility traces from the largest online map service in China. We conduct extensive experiments for evaluation. The results demonstrate that C-Watcher can accurately predict the potential risk of massive residential neighborhoods in a large number of Chinese cities. With large datasets, C-Watcher makes insightful suggestions on preventing the epidemic of COVID-19 alike for different residential neighborhoods via feature importance. 
\end{itemize}

\section{Notations and Related Work}
In this section, we first introduce the basic notations used throughout this paper, and then we formally formulate the research problem for early detection of high-risk neighborhoods. Last, we review the studies that are relevant to our work with the most related work discussed.

\paragraph{Notations and Formulation.}
We use $\bm{n}$ to denote the features of a residential neighborhood which will be presented in Section \ref{sec:feature}, and use $y$ to denote the binary label meaning whether the neighborhood is high risky ($y=1$) or not ($y=0$). The detection problem can be defined as: $f(\bm{n}) \rightarrow y$ where the function $f(\cdot)$ can be any machine learning model like Multi-Layer Perceptron (MLP).

The objective of C-Watcher is to make early detection of high-risk neighborhoods without epidemic outbreaks. Instead of relying on the confirmed infection cases to make a prediction which deemed to be time-delayed like  \cite{fu2020aware}, we assume that the COVID-19 epidemic only outbroke in epicenter cities (such as Wuhan in China) and no prior knowledge of confirmed cases, spreading trend or known hazard neighborhoods in target cities can be referred to. Such a cross-city prediction problem of latent high risky residential neighborhoods can be formulated as:
\begin{equation}
\label{eq-1}
    f_{cross}(\bm{n}^{T}|\,\{(\bm{n}_i^{E}\!, y_i^{E})\}) \rightarrow y^{T}
\end{equation}
where $\bm{n}^T$ and $y^T$ denote the features and binary label of a residential neighborhood in the target city. $\bm{n}_i^E$ and $y_i^E$ denote the features and label of a neighborhood from epicenter cities set. Hereafter, we omit subscript $i$ for simplicity. The $f_{cross}(\cdot)$ is a cross-city transfer learning model which is trained without ground-truth information in the target city.

\paragraph{Related Work.}
Aiming to fight against the COVID-19 pandemic, researchers in the computer science community carried out many studies from several perspectives recently. For instance, \citet{huang2020kddtransp} exhibit that user transportation-related behaviors in China have indeed been impacted by the containment measures during the COVID-19 pandemic. 
There are also a few studies \cite{huang2020quantifying,xiong2020understanding,liu2020Investigation} investigating the human mobility, the local economy, and the information acquisition during the COVID-19 outbreak in China while most of these studies remain at city level.

Some studies also demonstrate the effectiveness of mobility data for controlling the spread of COVID-19.
 \citet{vollmer2020using} exploit a Bayesian semi-mechanism model with mobility data to show the effectiveness to slow down the spread of the virus by constraints on individual movements and social interactions.
Based on the integration of mobility data and the global epidemic model \cite{balcan2009multiscale}, a study also reveals the effectiveness of fine-grained targeted mobility control policies towards the COVID-19 pandemic \cite{hao2020understanding}. The mobility data can also be integrated with compartmental models in epidemiology (like Susceptible-Exposed-Infected-Recovered (SEIR) model) \cite{ghamizi2020data} to better predict the epidemic dynamics.

\paragraph{Discussion.}
From the problems and methodologies perspectives, the most relevant work to our study includes~\cite{fu2020aware} and~\cite{xu2019deep,peng2019cm,mai2020modality}. Compared to~\cite{fu2020aware}, which smooths the confirmed cases of infections over spatial domains and predicts hazard areas during the COVID-19 outbreaks using simple spatial features like distance, C-Watcher system tackles the time delay and coarse-grained granularity issues and can early detect the high-risk  residential neighborhoods even before the outbreaks, through leveraging features derived from long-term/regular human mobility patterns. In terms of methodologies, though a great number of algorithms have been proposed for adversarial representation learning~\citet{makhzani2015adversarial}, adversarial metric learning~\cite{xu2019deep} and cross-modalities~\cite{mai2020modality}, our work is the first to study the city-invariant representation learning through Generative Adversarial Networks \cite{goodfellow2014generative} in the context of urban computing and COVID-19 prediction.

\section{Features for Neighborhood Detection}\label{sec:feature}
In this section, we present how to construct features from mobility data to characterize a residential neighborhood for early risk detection. We first introduce the data source used in our framework, and then three groups of constructed features are briefly discussed which are Point of Interest (POI) radius features (see Section \ref{sec:poi_radius}), demographic features (see Section \ref{sec:demographic features}) and transportation-related features (see Section \ref{sec:Transport-related features}), respectively. More details about the feature construction can be found in the Appendix \ref{apx:feacon}. 

The feature construction is mainly based on three data sources: POI basic property data, user profile data and human mobility data. 
POI basic property data contains the basic information of a POI, such as name, coordinates and types, which provides many semantic information for a POI \cite{10.1145/3394486.3403318,yuan2020spatio,hu2019we}. This data enable us to analyze the spatial relationship between neighborhoods and different types of POIs, such as hospitals, schools and bus stops \cite{Li2020competitive}. 
The user profile data are obtained from a user profile platform of Baidu which can return profile features for almost all internet users in China, such as gender, age and educational level. 
Human mobility data, collected from Baidu Maps in China, record search and transportation behaviors of the map users.

\subsection{POI Radius Features}\label{sec:poi_radius}
Here we introduce how to compute a group of POI radius features for a residential neighborhood based on POI basic property data. 
The intuition for this feature group is that basic living facilities around a residential neighborhood may have a correlation with the probability of its residents being infected by COVID-19.
For example, a neighborhood lacking basic living facilities may face a high risk, for the residents may passively go further away for basic living needs and face greater infection risks. Moreover, neighborhoods with poor living facilities often lack good property management, which may also lead to high infection risks.
To describe these living facilities related characteristics, we construct 15 POI radius features.  Each of them is defined as the shortest distance between the neighborhood and one certain type of POIs. All the used types of POIs are listed in the Appendix \ref{apx:feacon}. 

\begin{figure}
  \centering
    \subfigure[]{
    \label{infrastructure}
    \includegraphics[width=0.48\columnwidth]{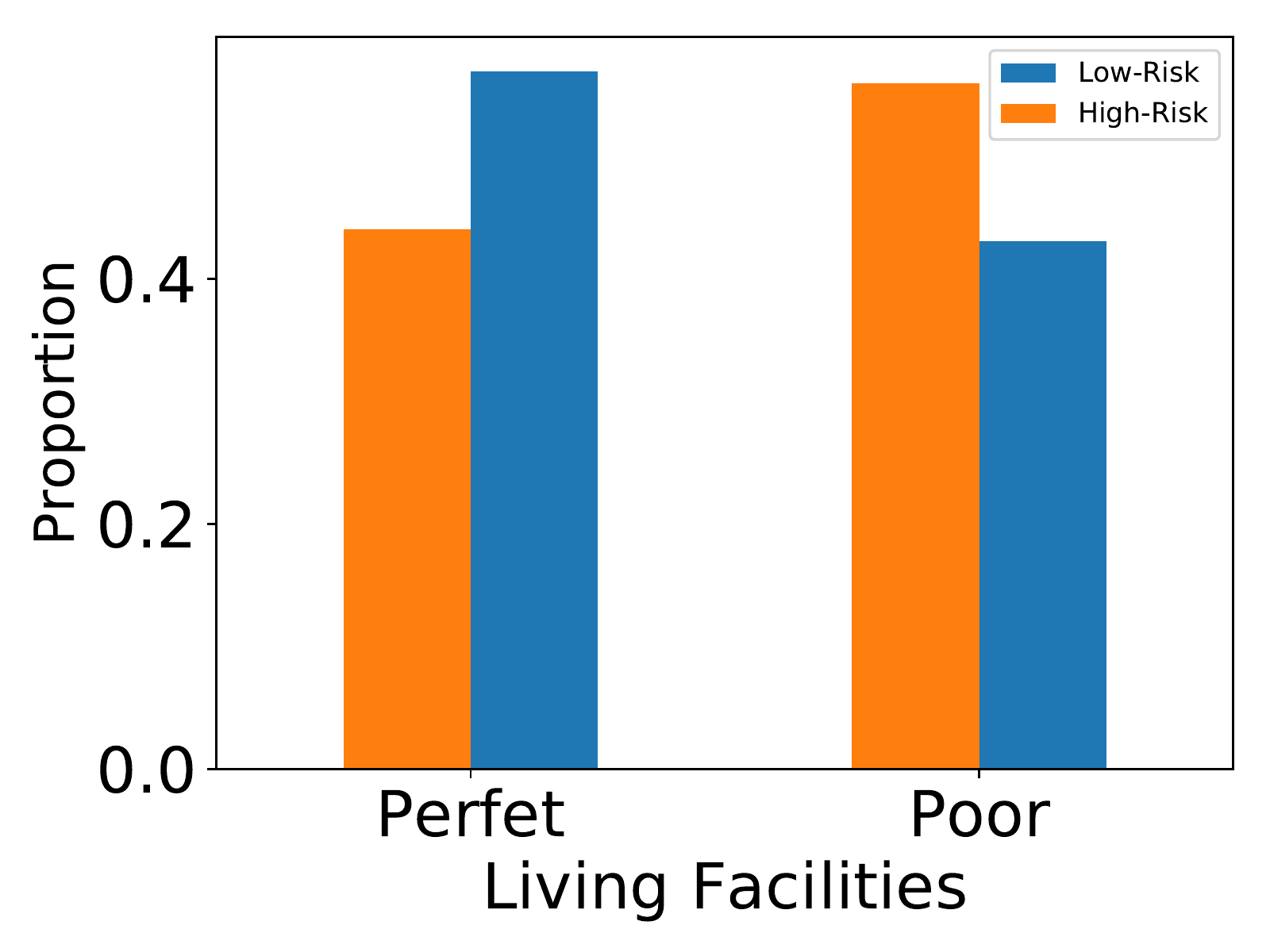}}
     \hspace{-1mm}
    \subfigure[]{
    \label{population_density}
    \includegraphics[width=0.48\columnwidth]{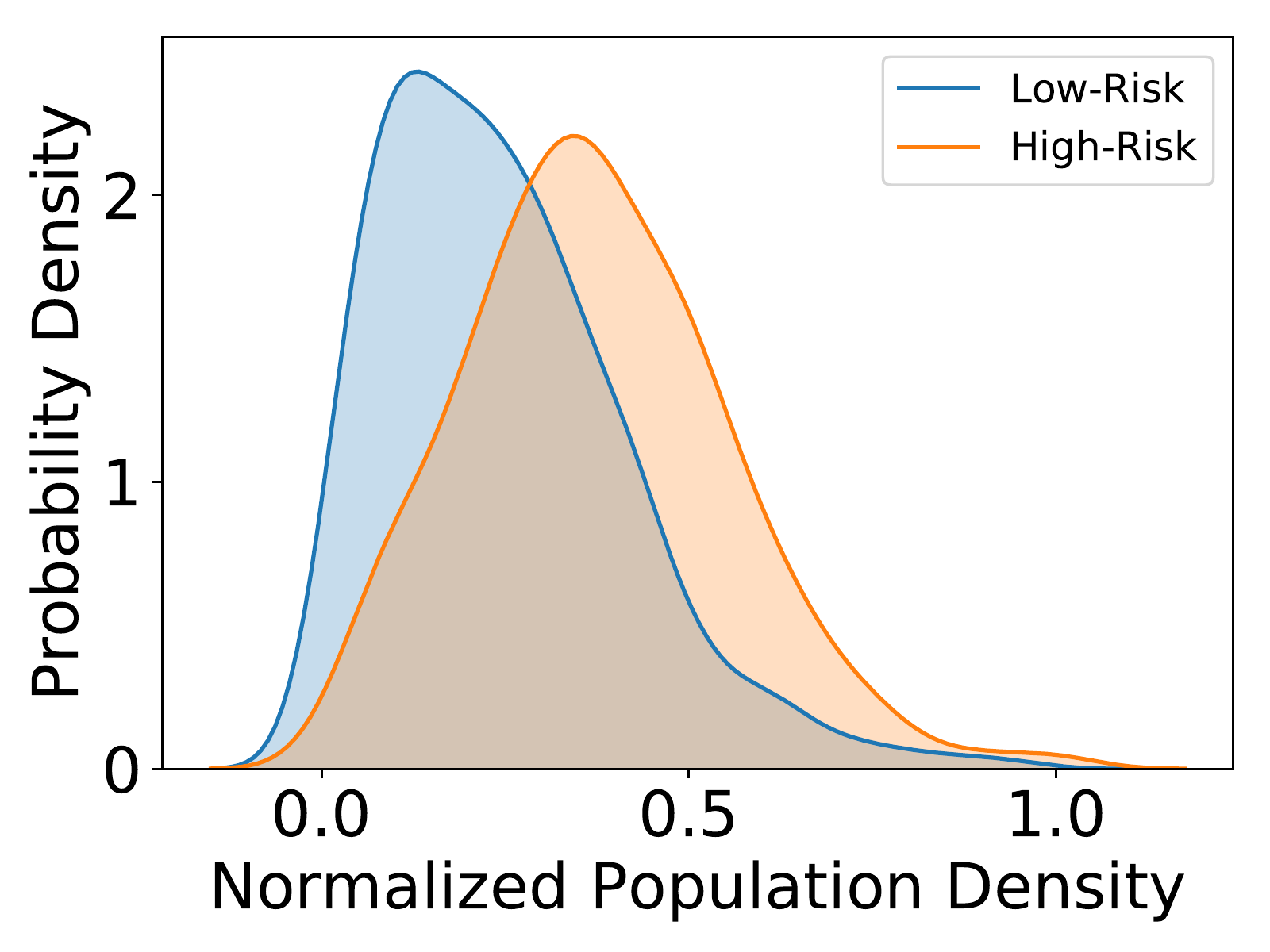}}
  \caption{Features of living facilities and population density visual analysis.}
  \label{feature_distribution} 
\end{figure}

Meanwhile, we define an additional binary feature to directly represent the perfect degree of living facilities. The value of this feature will be assigned as ``perfect'' if a set of basic living facilities (e.g. hospital, bus stop and so on) are all within $1 km$ of the given neighborhood. Otherwise, it is assigned ``poor''. 
The list of basic living facilities is also shown in the Appendix \ref{apx:feacon}.
We collect the high-risk and low-risk neighborhoods data in Wuhan city in China which is officially announced by the local government. 
Figure \ref{infrastructure} presents the ratio distribution of high-risk and low-risk neighborhoods grouping by this ``perfect-poor'' facility label in Wuhan data.
As we can see from Figure \ref{infrastructure}, for the neighborhoods with feature value as ``perfect'', the ratio of low-risk neighborhoods and high-risk ones is $0.57:0.44$; whereas the ratio of them for ``poor'' ones is $0.43:0.56$. It indicates that more high-risk neighborhoods have poor living facilities, while the low-risk neighborhoods are just the opposite.

\subsection{Demographic Features}\label{sec:demographic features}
Next, we present the demographic features of a residential neighborhood. At first, given that the COVID-19 is easy to transmit in a person-to-person way \cite{liu2020aerodynamic}, it is necessary to take into account population density for infection risks prediction. As Figure  \ref{population_density} illustrates, on average high-risk neighborhoods do have a higher population density than low-risk neighborhoods in Wuhan city. We also compute average commute distance as a feature for each neighborhood since residents with long commutes have high infection risks.

Moreover, different groups of residents may face different risk levels in a neighborhood. For example, old people and children are easier to be infected. And residents with higher educational levels may pay more attention to scientific prevention. Hence, we construct 11 features based on the distribution of residents according to different human attributes. We present each of these features as a vector of histogram statistics of residents' distribution. The full list of such attributes is provided in the Appendix \ref{apx:feacon}.  

\subsection{Transportation-Related Features}\label{sec:Transport-related features}
We also extract features of transportation-related behaviors from human mobility data to help predict infection risks. There have been some studies to prove that transportation-related behaviors have a close relationship with COVID-19 contagion spreading \cite{huang2020kddtransp}.  The transportation-related behaviors typically are recognized as the origin-transportation-destination (OTD) information\cite{8708931,xu2016taxi,Polestar}. Thus, we consider detailed features from the perspectives of T (transportation), OD (origin \& destination venues) and OTD (origin-transportation-destination pattern). All the features are extracted from the search and transportation data of Baidu Maps in a certain time period. Previous studies have shown that map search behavior is a leading indicator and predictor for crowd dynamics  \cite{zhou2018early}.

A vector in which the value of each element equals the corresponding ratio of transportation means, mainly including walk, bicycle, public transit and private vehicle, is used to depict the ``T feature'' for a residential neighborhood. 
The ``OD feature''  consists of types of visit venues and the distance between origin and destination venues.
We classify the destination venues according to their types (e.g. hospital, restaurant, hotel, and school) and compute the proportion of each type. We also extract origin-destination distance and categorize it into different distance buckets. The proportions of different distance buckets for the neighborhood are also formed as a feature vector. Moreover, since the OTD (origin-transportation-destination) patterns most directly reflect human mobility, we collect the top-20 hottest travel patterns from all the cities in our dataset, which is a triplet tuple composed of the type of origin venue (residential area), means of transportation and type of destination venue. The histogram distribution of these top-20 OTD travel patterns of each neighborhood is treated as ``OTD'' features. More details about the transportation-related features can be found in the Appendix \ref{apx:feacon}. 

\section{Cross-City Transfer Learning}\label{sec:transer frame work}
\begin{figure}[t]
\centering
\includegraphics[width=0.48\textwidth]{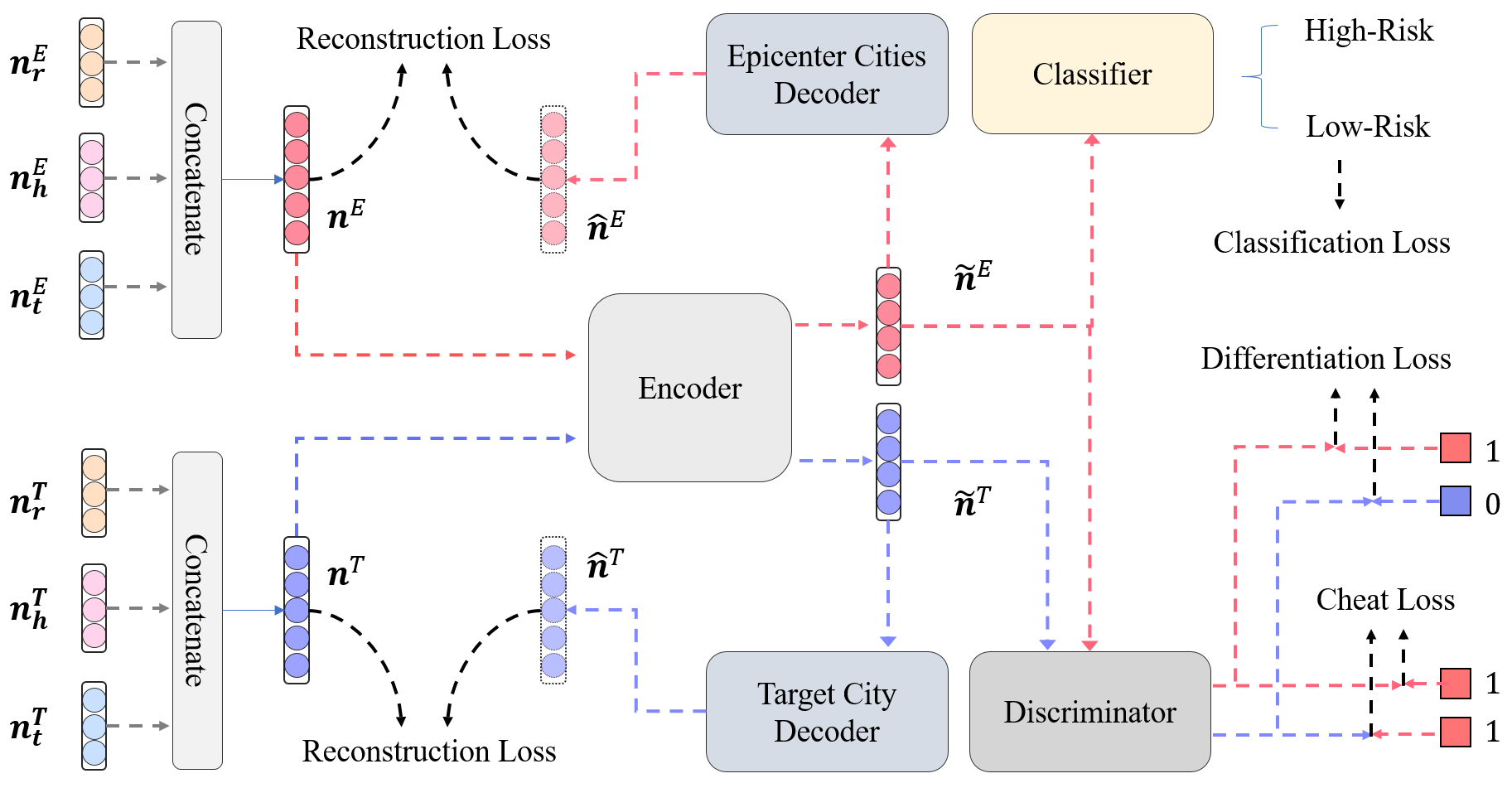} %
\caption{Illustration of cross-city transfer learning model of C-Watcher.}
\label{model-frame}
\end{figure}

In this section, we present the cross-city transfer learning model, which is a core component of C-Watcher, to improve the performance of early detection of high-risk neighborhoods via transferring the knowledge about COVID-19 infection from the epicenter to the target city. 
Usually, discrepancies always exist between different cities. Thus, we intend to learn city-invariant knowledge applicable to both epicenters and target cities, instead of those characteristics unique to epicenter cities. 

\subsection{Overview}\label{sec:framework overview}
An overview of our proposed cross-city transfer learning model with four components is given in Figure \ref{model-frame}. 
The first component is a neural network encoder used to learn the representation of a neighborhood on the basis of three groups of features introduced in Section \ref{sec:feature}. However,
the discrepancy of input distributions in different cities may lead to the gap between encoded representations of epicenter cities and target cities, which may severely disrupt the detection ability in target cities. 
Thus, we adopt adversarial learning by adding a discriminator component to identify whether the output of the encoder belongs to the target city or not. In addition, to ensure that the embedded feature of the encoder still keeps the ability to depict the residential neighborhoods, we exert two decoders which recover features of epicenter cities and target city respectively from the output of the encoder. Moreover, to achieve the prediction goal of C-Watcher, a classifier is also added to optimize the learned representations space and make it more related to the prediction of COVID-19 infection risks.

In the following sections, we present our cross-city transfer learning model in detail by introducing how each component works. Here comes the notation of model input first. For both residential neighborhoods in epicenter cities and target cities, we have a feature vector consist of three groups:
\begin{equation}
\begin{split}
\label{eq-2}
    \bm{n}^{E} &= cat(\bm{n}_{r}^{E}, \bm{n}_{h}^{E}, \bm{n}_{t}^{E})\\
    \bm{n}^{T} &= cat(\bm{n}_{r}^{T}, \bm{n}_{h}^{T}, \bm{n}_{t}^{T})
\end{split}
\end{equation}
where $\bm{n}^E$ denotes all of the features of a residential neighborhood in epicenter cities. $\bm{n}_r^E$, $\bm{n}_h^E$ and $\bm{n}_t^E$ respectively denote the POI radius features, demographic features and transportation-related features of that residential neighborhood in epicenter cities. $\bm{n}^T$, $\bm{n}_r^T$, $\bm{n}_h^T$ and $\bm{n}_t^T$ similarly denote the corresponding features of residential ones in target city. The function $cat(\cdot)$ is the concatenating operation.

\subsection{City-Invariant Representation Learning}\label{asversarial learning}
Since learning unique characteristics of neighborhoods in epicenter cities brings little benefit for early risk detection in target cities,  we propose a city-invariant representation learning method, which is inspired by the multi-mode adversarial representation learning methods \cite{mai2020modality,makhzani2015adversarial}.  Here the encoder is a transformer of data distribution. Given the input vectors $\bm{n}^E$ and $\bm{n}^T$, we use $\widetilde{\bm{n}}^E$ and $\widetilde{\bm{n}}^T$ to denote the outputs of the encoder. Similar to \cite{mai2020modality} and \cite{makhzani2015adversarial}, the distributions transformation from the inputs to encoded representations can be presented as:
\begin{equation}
\begin{split}
\label{eq-3}
    p(\widetilde{\bm{n}}^E,\Phi_e) &= \int_{\bm{n}^E} e(\widetilde{\bm{n}}^E |\,\bm{n}^E, \Phi_e)\,p(\bm{n}^E)\,d\bm{n}^E\\
    p(\widetilde{\bm{n}}^T,\Phi_e) &= \int_{\bm{n}^T} e(\widetilde{\bm{n}}^T |\,\bm{n}^T, \Phi_e)\,p(\bm{n}^T)\,d\bm{n}^T
\end{split} 
\end{equation}
where $p(\cdot)$ denotes the data distribution and $e(\cdot \,,\Phi_e)$ represents the encoding distribution. $\Phi_e$ are parameters of the encoder, determining the projection space where the distributions of input data $p(\bm{n}^E)$ and $p(\bm{n}^T)$ are transformed into that of encoded representations $p(\widetilde{\bm{n}}^E,\Phi_e)$ and $p(\widetilde{\bm{n}}^T,\Phi_e)$. 

In general, $p(\widetilde{\bm{n}}^E,\Phi_e)$ and $p(\widetilde{\bm{n}}^T,\Phi_e)$ are different distributions characterizing different cities.
It means that some features helpful in predicting COVID-19 infection risk may be unique to neighborhoods in epicenter cities unless we impose constraints on the encoder. To this end, we use adversarial learning to narrow the discrepancies between distributions $p(\widetilde{\bm{n}}^E,\Phi_e)$ and $p(\widetilde{\bm{n}}^T,\Phi_e)$ by adding a discriminator to distinguish whether the neighborhood comes from epicenter cities or the target city. 
In this way, the discriminator needs to do a binary classification task, in which it takes the encoded representations as inputs and aims to identify the inputs $\widetilde{\bm{n}}^E$ from epicenter cities as true but the inputs $\widetilde{\bm{n}}^T$ from target cities as false, while the encoder tries its best to confuse the discriminator to classify both of them as true.
We can formulate the function of discriminator as:
\begin{equation}
\begin{split}
\label{eq-4}
    D(\widetilde{\bm{n}}^E,\Phi_D) \rightarrow 1\\
    D(\widetilde{\bm{n}}^T,\Phi_D) \rightarrow 0
\end{split}
\end{equation}
where $D(\,\cdot,\Phi_D)$ denotes the function of the discriminator which can be an MLP model that outputs the probability from 0 to 1. On the contrary, the encoder competes against the discriminator by:
\begin{equation}
\begin{split}
\label{eq-5}
    D(\widetilde{\bm{n}}^E,\Phi_D) \rightarrow 1\\
    D(\widetilde{\bm{n}}^T,\Phi_D) \rightarrow 1
\end{split}
\end{equation}

For this adversarial learning procedure, we use binary cross entropy (BCE) to define the loss function:
\begin{align}
\label{eq-6}
\mathcal{L}_{al} &= \mathcal{L}_{diff}(\widetilde{\bm{n}}^E,\widetilde{\bm{n}}^T) + \mathcal{L}_{ch}(\widetilde{\bm{n}}^E,\widetilde{\bm{n}}^T)\\
\mathcal{L}_{diff} &= -[log(D(\widetilde{\bm{n}}^E)) + 
log(1-D(\widetilde{\bm{n}}^T)]\\
\mathcal{L}_{ch} &= -[log(D(\widetilde{\bm{n}}^E)) + 
log(D(\widetilde{\bm{n}}^T)]
\end{align}
where $D(\widetilde{\bm{n}}^E)$ is used to represent $D(\widetilde{\bm{n}}^E, \Phi_D)$ in simplicity and so does $D(\widetilde{\bm{n}}^T)$. The differentiation loss $\mathcal{L}_{diff}$ guides discriminator to predict $\widetilde{\bm{n}}^E$ as true (epicenter cities) but $\widetilde{\bm{n}}^T$ as false (target city), while the encoder tries to learn features that are common between epicenter cities and target city to hinder discriminator from distinguishing successfully, under the effects of cheat loss $\mathcal{L}_{ch}$.
The adversarial procedure will finally reach an equilibrium situation where the discriminator could no longer distinguish whether the encoded representations come from epicenter cities or target city, then the encoder is able to extract ``city-invariant'' features from raw inputs $\bm{n}^E$ and $\bm{n}^T$.
In this case, discrepancies between cities decrease and the experience which helps predict infection risks in epicenter cities can make more sense in target cities.

\subsection{Embedding Space Constraints}\label{space constrains}
A problem about city-invariant representation learning is that, if no regulations and restrictions are imposed on the embedding space of the encoder, the encoded representations of epicenter cities and target cities may only be similar in distribution but fail to retain useful information for identifying high-risk neighborhoods. 
We solve this problem with multi-task learning strategy by 
additionally exerting an auto encoder-decoder features reconstruction component, as well as a COVID-19 infection risks prediction component.

The reconstruction component consists of two decoders (one for residential neighborhoods in epicenter cities and another one for residential neighborhoods in the target city) which take the encoded representations as inputs. The decoding operation can also be considered as a distribution transformation like encoding:
\begin{equation}
\begin{split}
\label{eq-9}
    p(\widehat{\bm{n}}^E,\Phi_d^E) &= \int_{\widetilde{\bm{n}}^E} d^E(\widehat{\bm{n}}^E |\,\widetilde{\bm{n}}^E, \Phi_d^E)\,p(\widetilde{\bm{n}}^E)\,d\widetilde{\bm{n}}^E\\
    p(\widehat{\bm{n}}^T,\Phi_d^T) &= \int_{\widetilde{\bm{n}}^T} d^T(\widehat{\bm{n}}^T |\,\widetilde{\bm{n}}^T, \Phi_d^T)\,p(\widetilde{\bm{n}}^T)\,d\widetilde{\bm{n}}^T
\end{split}
\end{equation}
where $\widehat{\bm{n}}^E$, $\widehat{\bm{n}}^T$ denote the reconstructed outputs of decoders from $\widetilde{\bm{n}}^E$ and $\widetilde{\bm{n}}^T$ respectively, and $d^E(\widehat{\bm{n}}^E |\,\widetilde{\bm{n}}^E,\Phi_d^E)$ represents the epicenter cities decoder function with parameters $\Phi_d^E$, while $d^T(\widehat{\bm{n}}^T |\,\widetilde{\bm{n}}^T,\Phi_d^T)$ is similar but for the target city. Aiming to approximate decoded representations to the original inputs ($\widehat{\bm{n}}^E \rightarrow \bm{n}^E$ and $\widehat{\bm{n}}^T \rightarrow \bm{n}^T$), we use mean square error to define reconstruction loss function:
\begin{align}
\label{eq-10}
    \mathcal{L}_{rec} &= \mathcal{L}_{rec}^E + \mathcal{L}_{rec}^T\\
    &= \|\,\widehat{\bm{n}}^E,\bm{n}^E\|_2 + \|\,\widehat{\bm{n}}^T,\bm{n}^T\|_2
\end{align}
Optimized by the reconstruction loss above, the encoder-decoder framework ensures that the embedding space is still characterizing a residential neighborhood.

Moreover, considering that our ultimate objective is to detect latent high-risk neighborhoods, we add a classifier to identify COVID-19 infection risks in epicenter cities upon the learned encoded representations. The classification problem can be defined as :
\begin{equation}
\label{eq-11}
    C(\widetilde{\bm{n}}^E,\Phi_c) \rightarrow y^E,\,  y^E \in \{0,1\}
\end{equation}
where $C(\cdot,\Phi_c)$ denotes the function of MLP classifier with parameter $\Phi_c$. This is also a binary classification task and we use BCE to define the classification loss function:
\begin{equation}
\label{eq-12}
    \mathcal{L}_{cl} \!=\! -y^{E}\!log(C(\widetilde{\bm{n}}^E\!\!,\Phi_c)) \!-\! 
    (1 \!-\! y^{E})log(1 \!-\! C(\widetilde{\bm{n}}^E\!\!,\Phi_c))
\end{equation}
The classification loss transmits the known information carried by label $y^E$  to encoder and classifier, which is COVID-19 infection risks of neighborhoods in epicenter cities. It achieves the goals to restrict the encoded representations to be instructive in high-risk neighborhood identification.

All in all, loss functions generated from all the three components of discriminator, decoders and classifier will act on the encoder and optimize the embedding space in our proposed cross-city transfer model. The total loss function can be expressed qualitatively as:
\begin{equation}
\label{eq-13}
    \mathcal{L} = \lambda_{diff}\mathcal{L}_{diff} + \lambda_{ch}\mathcal{L}_{ch} + \lambda_{rec}\mathcal{L}_{rec} + \lambda_{cl}\mathcal{L}_{cl}
\end{equation}
In model training, the adversarial model is optimized in an alternate mode. We use differentiation loss $\mathcal{L}_{diff}$ to optimize the discriminator first to improve its discriminatory ability to neighborhoods in both epicenter cities and that target city, which also leads to the rise of cheat loss. Then we apply cheat loss $\mathcal{L}_{ch}$ combined with $\mathcal{L}_{rec}$ and $\mathcal{L}_{cl}$, to guide the encoder to optimize its parameters in 
a direction where demands to learn city-invariant, informative and risk-discriminative features are all taken into consideration. Together with the encoder, the decoders and classifier update themselves based on $\mathcal{L}_{rec}$ and $\mathcal{L}_{cl}$, respectively.

\subsubsection{Reference City Validation Mechanism.}\label{sec: anchor}
\begin{figure}[t]
\centering
\includegraphics[width=0.3\textwidth]{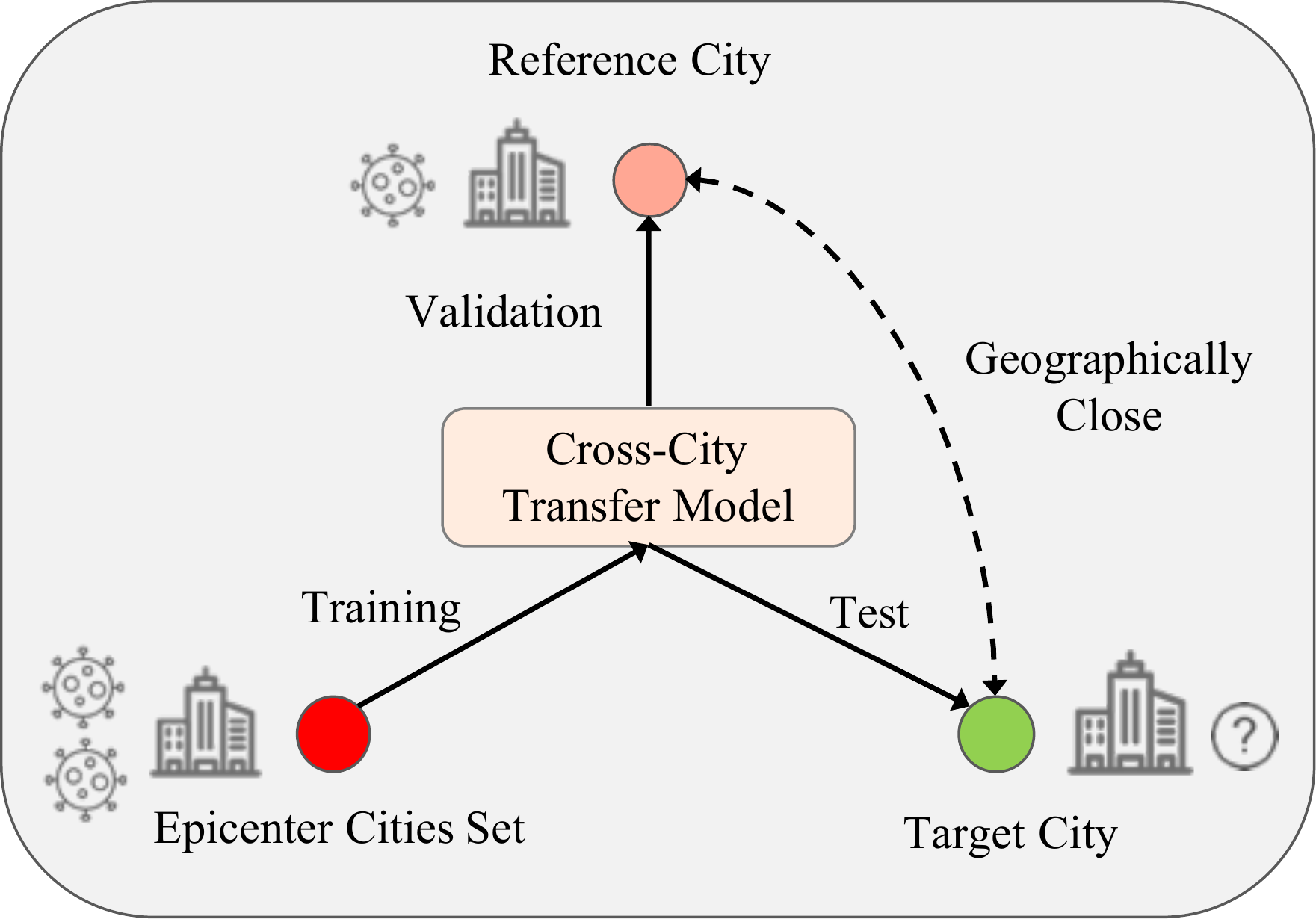}
\caption{Diagram of reference city validation mechanism.}
\label{anchor-frame}
\end{figure}
Another problem of C-Watcher is how to select the best hyperparameters to train the model. Here we build a reference city validation mechanism to tune hyperparameters. The illustrated diagram is shown in Figure \ref{anchor-frame}. Reference city in our paper can be  epicenter cities, and can also be some cities with COVID-19 outbreak but not so serious as epicenters. We train the C-Watcher model on epicenter cities set, and use ground truth data of the reference city as validation data to choose the hyperparameters.
Then we evaluate the early detection performance in target cities geographically close to that reference city.
In this case, we ensure that our trained model to detect latent high-risk neighborhoods in a target city with best hyperparameters, without any prior information related to COVID-19 confirmed cases and spreading trend.

\section{Experiments}\label{sec:experiments}

\subsection{Datasets and Settings}\label{sec:datasets and settings}
\subsubsection{Dataset construction.}\label{sec: dataset construction}
The constructed datasets simulate a common outbreak pattern in a country. In the scenarios, there is a set of epicenter cities in the country (like Wuhan in China), and a few reference cities (see Section \ref{sec: anchor}) which have some confirmed cases. The C-Watcher can be trained on the epicenter cities and reference cities datasets, and then be used to make early detection of high-risk residential neighborhoods in the rest of cities in the country.\footnote{The code can be found at \url{https://github.com/PaddlePaddle/Research/tree/master/ST_DM/AAAI2021-CWatcher/}.}


In this evaluation, all the datasets are built based on 16 cities in China which consists of one dataset from epicenter city, 5 evaluation datasets from selected reference cities, and 10 test datasets from other cities. The epicenter city dataset is constructed based on Wuhan, which has the largest number of confirmed cases in China and is well-recognized as the epicenter of the COVID-19 
outbreak in China. 
The five selected reference cities, 
Shenzhen, Changsha, Chengdu, Shanghai and Zhengzhou, are key cities in their provinces and they are also evenly situated in different geographical regions of China. For each reference city, we also construct two test datasets from two cities geographically closed to them. 
The full list of test cities is in the Appendix \ref{apx:sod}. 
The POI data and user profile data of all the cities are both collected by the first week of March 2020. The human mobility data are collected from January 1, 2020 to March 3, 2020.

We also make a great effort to build the ground-truth dataset. For the Wuhan dataset, we manually collected all the high-risk residential neighborhoods (released on February 24, 2020) and low-risk residential neighborhoods (released on March 6, 2020) which are officially published by the local government. 
After data cleaning and feature alignment, there are 336 high-risk neighborhoods and 715 low-risk neighborhoods. The statistics of high-risk neighborhoods in other cities datasets are listed in Appendix \ref{apx:sod}. 
For datasets of other cities, we label the neighborhoods with at least one confirmed case as high-risk while others as low-risk, based on the public COVID-19 patients dataset by \cite{fu2020aware}.

In order to tune hyperparameters for baselines, we split Wuhan dataset into three folds as train, validation and test data by a 0.7:0.15:0.15 ratio. The hyperparameters tuning for C-Watcher is done by reference city validation mechanism (see Section \ref{sec: anchor}). 

\subsubsection{Baselines.}\label{sec:baseline}
Since we are the first to study the COVID-19 high-risk neighborhoods early detection problem, there is no direct competitor of C-Watcher. Thus, we compare  C-Watcher with classical machine learning methods of Multi-Layer Perceptron (MLP), Support Vector Machine (SVM), XGBoost (XGB) and Lasso Logistic Regression (Lasso-R). We use the dataset from epicenter city to train the baselines, and make a prediction on the datasets of test cities.

\subsubsection{Metrics.}\label{sec:metrics}
Since the detection of high-risk neighborhoods is an imbalance binary classification task (high-risk neighborhoods are much less than low-risk ones), we mainly evaluate the performance by AUC (Area under the ROC Curve), which reflects model performance within different discrimination thresholds \cite{manning2008introduction, fu2020aware}. In addition, we also calculate the p-value by pairwise t-test between baselines and C-Watcher to show the statistical significance of the evaluation results. 

\subsubsection{Optimization and hyperparameters tuning.} \label{sec: hyperparameters}
We optimize C-Watcher by Adam optimizer. The main hyperparameters of C-Watcher, including weights of the loss function ($\lambda_{ch}$, $\lambda_{diff}$, $\lambda_{rec}$ and $\lambda_{cl}$ ), learning rate and hidden size of the neural network of each component are determined by grid search method, with batch size fixed as 64. 

\subsection{Performance Evaluation of Early Detection}\label{sec: main-exp}
\begin{table*}[t]
    \centering
	\begin{tabular}{c|c|c|c|c|c|c|c}
		\toprule
    	\multirow{2}{*}{} & \multicolumn{2}{c|}{Overall} & Huizhou & Shaoyang & Lianyungang & Xuchang & Chongqing\\
        \cmidrule{2-8}
    	 & AUC & P-value & \multicolumn{5}{c}{AUC} \\
		\midrule
		SVM & 0.5999 & 0.0005 & 0.7049 & 0.5615 & 0.6728 & \textbf{0.7330} & 0.5693 \\
		\hline
		\rule{0pt}{10pt}
		XGB & 0.5810 & 0.0018 & 0.6266 & 0.5190 & 0.6182 & 0.7067 & 0.4901 \\
		\hline
		\rule{0pt}{10pt}
		Lasso-R & 0.5853 & 0.0006 & 0.6364 & 0.5410 & 0.6515 & 0.7195 & 0.5718 \\
		\hline
		\rule{0pt}{10pt}
		MLP & 0.5963 & 0.0005 & 0.6995 & 0.5594 & 0.6850 & 0.7278 & 0.5438 \\
		\hline
		\rule{0pt}{10pt}
		C-Watcher & \textbf{0.6490} & -- & \textbf{0.7352} & \textbf{0.6433} & \textbf{0.7218} & 0.7312 & \textbf{0.6142} \\
		\bottomrule
	\end{tabular}
	\caption{Early detection performance comparison between C-Watcher and baselines on cross-city datasets. The ``target city - reference city'' relationship are ``Huizhou - Shenzhen'', ``Shaoyang - Changsha'', ``Lianyungang - Shanghai'', ``Xuchang - Zhengzhou'' and `` Chongqing - Chengdu''.}
	\label{table_main_exp}
\end{table*}
We evaluate the performance of C-Watcher and its baselines for early detection of high-risk neighborhoods on test datasets of the 10 target cities. 
In Table \ref{table_main_exp}, the overall column shows the average AUC of the 10 cities.  
We can see that C-Watcher can improve the AUC by 8.18\% over the best baseline (SVM and MLP). We also conduct pairwise t-test between the C-Watcher and each baseline. The p-values in 
Table \ref{table_main_exp} demonstrate that C-Watcher can achieve significantly better performance than other baselines.

We also show the prediction performance on test datasets of five target cities in Table \ref{table_main_exp}. Each target city corresponds to one reference city. We can see that the improvement by C-Watcher over baselines in different cities is different. For example, the improvement by C-Watcher over the best baseline on Shaoyang is 14.57\% (i.e., C-Watcher (0.6433) vs. SVM (0.5615)); but the one by C-Watcher over the best baseline on Xuchang is about 0\% (the AUC of C-Watcher is almost the same with other baselines). It is an interesting problem to investigate what factors impact the performance of transfer learning of C-Watcher. A possible reason is that some geographically closed cities are not similar, thus the reference city cannot help to select the best hyperparameters for transfer learning.  We leave this problem as a further research investigation.  We put the prediction performance of all the ten cities in the Appendix \ref{apx:red}.  

\subsection{Feature Importance}\label{sec:feature importance}
\begin{figure}[b!]
\centering
\includegraphics[width=0.45\textwidth]{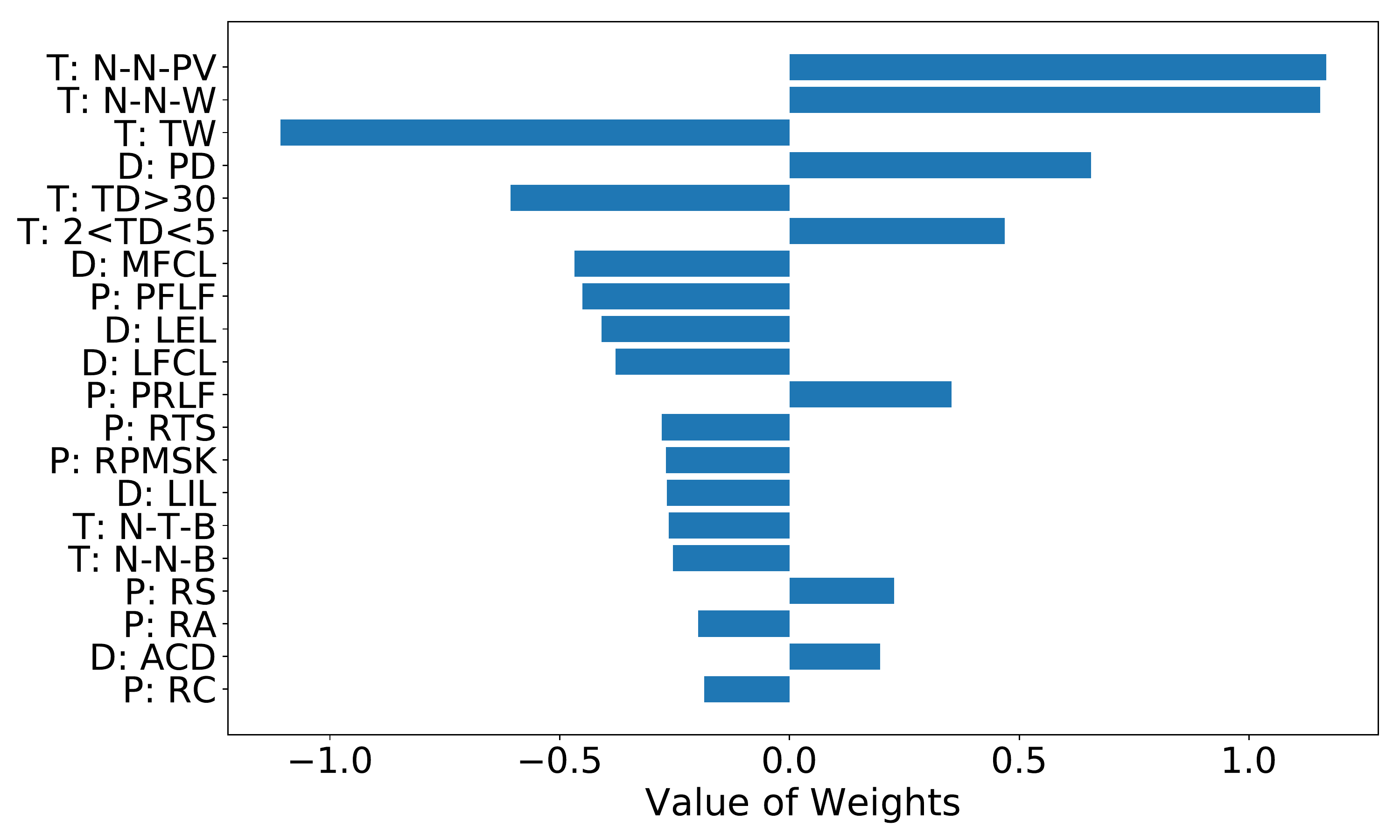}
\caption{The top-20 most important features for high-risk neighborhoods detection.}
\label{features-importances}
\end{figure}
Here we conduct a feature importance analysis to discuss possible characteristics of neighborhoods leading to the high risk for infection. We use Lasso Logistic Regression (Lasso-R) on epicenter Wuhan dataset to select the top-20 important features according to the absolute coefficient value, which is illustrated in Figure \ref{features-importances}. The full name of each feature is listed in the Appendix \ref{apx:fafn}. 
The feature importance analysis reveals several insightful and interesting points for preventing the epidemic. For POI radius features, except the effect of perfect and poor living facility of a neighborhood (which are denoted by ``P:PFLF'' and ``P:PRLF'' in Figure \ref{features-importances}), the coefficient of ``P:RTS'' indicates that the long distance to a train station can reduce the risk of the neighborhood. For the demographic features, except the high population density (denoted by ``D:PD''), the long average commute distance (denoted by ``D:ACD'') also increases the risk of the neighborhood. For the transportation-related features, we find that the percentage of travelling on walk (denoted by ``T:TW'') can reduce the risk of the neighborhood by a large margin. 
We believe such analysis can help us identify factors for high-risk neighborhoods, and provide insightful suggestions on preventing the epidemic of COVID-19 in future.

\subsection{Effectiveness of  Feature Groups}\label{sec:feature effectiveness}
In this section, we verify the effectiveness of 3 groups of hand-crafted features. In specific, we separately evaluate the performance of each group of features in detecting high/low-risk neighborhoods, then we compare them with the performance of taking all 3 groups of features together as inputs. All the comparative experiments for feature effectiveness are conducted by MLP on the epicenter Wuhan dataset.
As we can see from Table \ref{table:feature effectiveness}, all the three groups of features can positively classify the high-risk and low-risk neighborhoods. More importantly, the combination of these three groups certainly improves the model's overall performance, which proves complementary among the three groups of features.
\begin{table}[htb]
    \centering
	\begin{tabular}{c|c}
		\hline
		\rule{0pt}{10pt}
		Feature groups & AUC \\
		\hline
		\rule{0pt}{10pt}
		POI Radius  & 0.8033 \\
		\hline
		\rule{0pt}{10pt}
		Demographic & 0.7579 \\
		\hline
		\rule{0pt}{10pt}
		Transportation-Related & 0.7414 \\
		\hline
		\rule{0pt}{10pt}
		All three Groups & \textbf{0.8458} \\
		\hline
	\end{tabular}
	\caption{Detection performance comparison of MLP with different feature groups on Wuhan dataset.}
	\label{table:feature effectiveness}
\end{table}

\section{Conclusion}
In this paper, we study the problem of predicting infection risks of COVID-19 in urban neighborhoods. We first construct a set of features incorporating human mobility data to characterize the demographic/socioeconomic status and spatial interactions of a residential neighborhood, then propose C-Watcher, a data-driven framework based on these features to early detect high-risk neighborhoods in a city ahead of local COVID-19 outbreaks. 
To improve infection risks identification in target cities, C-Watcher adopts adversarial learning algorithms that learn ``city-invariant'' features to boost generalizing knowledge witted in epicenter and build a reference city validation mechanism for hyperparameters selection. 
We conduct extensive experiments upon real-world data in the early stage of COVID-19 outbreaks from China to demonstrate the advantages of C-Watcher to early detect high-risk neighborhoods across cities and analyze the importance and effectiveness of explored features.

\section*{Acknowledgment}
We thank all reviewers for insightful comments.
This work is supported in part by National Key R\&D Program of China (No. 2018YFB1402600), and in part by grant from the National
Natural Science Foundation of China (Grant No. 91746301 and No. 71531001). Part of experiments in this paper was carried out using anonymous data and secure data analytics provided by Baidu Data Federation Platform (Baidu FedCube). For data accesses and usages, check http://fedcube.baidu.com/page\_en.html.
\bibliography{main}
\appendix
\section{Appendix}
\subsection{Feature Constructions}\label{apx:feacon}
\subsubsection{Types of POIs}
As mention in section \ref{sec:poi_radius}, we construct 15 POI radius features defined as the shortest distance between a neighborhood and one certain type POIs. The 15 types of POIs are as follows:
\emph{hospital}, \emph{clinic}, \emph{campus}, \emph{kindergarten \& primary school \& secondary school} (we see these three types as a whole), \emph{bus stop}, \emph{subway station}, \emph{airport} , \emph{train station}, \emph{coach station},  \emph{shopping mall}, \emph{supermarket}, \emph{market}, \emph{shop}, 
\emph{police station}, \emph{scenic spots}.

We also define another binary feature to directly reflect the perfect degree of basic living facilities and it will be assign “perfect” if there are all the following types of living facilities shown in Table \ref{table-apdx-facilities} within 1 km of the given neighborhood. The types of  living facilities within 1 km are mainly selected according to a official document released by Ministry of Housing and Urban-Rural Development of China \footnote{http://www.mohurd.gov.cn/wjfb/201811/W020181130044801\\.pdf}.

\subsubsection{Human Attributes Features}
In section \ref{sec:demographic features}, we introduce 11 demographic features of a neighborhood based on distributions of residents according to the attributes shown in Table \ref{table-apdx-profile}. 
\begin{table}[htbp]
    \centering
    \renewcommand\tabcolsep{5.0pt}
	\begin{tabular}{c|c||c|c}
		\toprule
		Feature & Attribute & Feature & Attribute \\
		\midrule
		\multirow{4}{*}{Age} & \textless 18 & \multirow{2}{*}{ Marriage} & Yes \\
		\cline{2-2}\cline{4-4}
		\rule{0pt}{10pt}
		 & 18-24 &  & No \\
		 \cline{2-4}
		\rule{0pt}{10pt}
		 & 25-34 &  & Low \\
		 \cline{2-2}\cline{4-4}
		\rule{0pt}{10pt}
		 & ...   &  Food & Middle\\
		 \cline{1-2}\cline{4-4}
		\rule{0pt}{10pt}
		 \multirow{2}{*}{Gender} & Male &  & High\\
		 \cline{2-4}
		 \rule{0pt}{10pt}
		 & Female & \multirow{2}{*}{Life Stage} & Student \\
		 \cline{1-2}\cline{4-4}
		 \rule{0pt}{10pt}
		 & \textless 2499 &  & ...\\
		 \cline{2-4}
		 \rule{0pt}{10pt}
		 Income & 2500-3999 &  & Manager \\
		 \cline{2-2}\cline{4-4}
		 \rule{0pt}{10pt}
		 & ... & Job & Technician\\
		 \cline{1-2}\cline{4-4}
		 \rule{0pt}{10pt}
		 & Low &  & ... \\
		 \cline{2-4}
		 \rule{0pt}{10pt}
		 Consumption & Middle &  & IT \\
		 \cline{2-2}\cline{4-4}
		 \rule{0pt}{10pt}
		 & High & Industry & Food \\
		 \cline{1-2}\cline{4-4}
		 \rule{0pt}{10pt}
		 & High School &  & ...\\
		 \cline{2-4}
		 \rule{0pt}{10pt}
		 Education & College & \multirow{2}{*}{Car} & Yes\\
		 \cline{2-2}\cline{4-4}
		 \rule{0pt}{10pt}
		 & Bachelor &  & No\\
		 \bottomrule
		
\end{tabular}
\caption{Features of human attributes.}
	\label{table-apdx-profile}
\end{table}

\subsubsection{Transportation-Related Features}
In section \ref{sec:Transport-related features}, we discuss transportation-related features from the perspectives of \textbf{T} (means of transportation) features, \textbf{OD} (types of visit venues and the distance between origin and destination venues) features and \textbf{OTD} (origin-transportation-destination pattern) features. The details of each feature are shown in Table \ref{table-apdx-OTD}.

\subsection{Statistics of Datasets}\label{apx:sod}
Table \ref{table-apdx-dataset} shows the statistics of high-risk neighborhoods of datasets used in our experiments. Note that the number of high-risk and low-risk neighborhoods of Wuhan is officially announced by local government. Thus, the total number of neighborhoods of Wuhan is small.
\begin{table}[h]
    \centering
	\begin{tabular}{c|c|c|c}
		\toprule
        \multicolumn{2}{c|}{Datasets} & High-Risk  & Total \\
		\hline
		\rule{0pt}{10pt}
		Epicenter City & Wuhan & 336 & 1051 \\
		\midrule
		 & Shenzhen & 852 & 28249 \\
		\cline{2-4}
		\rule{0pt}{10pt}
		 & Shanghai & 245 & 36852\\
		\cline{2-4}
		\rule{0pt}{10pt}
		Reference City & Changsha & 296 & 12145 \\
		\cline{2-4}
		\rule{0pt}{10pt}
		 & Zhengzhou & 155 & 14006\\
		\cline{2-4}
		\rule{0pt}{10pt}
		 & Chengdu & 198 & 27008\\
		\midrule
		 & Huizhou & 18 & 6403 \\
		\cline{2-4}
		\rule{0pt}{10pt}
		& Guangzhou & 659 & 26055 \\
		\cline{2-4}
		\rule{0pt}{10pt}
		 & Nanjing & 93 & 11854 \\
		\cline{2-4}
		\rule{0pt}{10pt}
		 & Lianyungang & 18 & 1774\\
		\cline{2-4}
		\rule{0pt}{10pt}
		\multirow{2}{*}{Target City} & Shaoyang & 26 & 1637 \\
		\cline{2-4}
		\rule{0pt}{10pt}
		 & Yiyang & 12 & 960 \\
		\cline{2-4}
		\rule{0pt}{10pt}
		 & Xuchang & 25 & 1799\\
		\cline{2-4}
		\rule{0pt}{10pt}
		 & Anyang & 9 & 875\\
		\cline{2-4}
		\rule{0pt}{10pt}
		 & Chongqing & 630 & 19137\\
		\cline{2-4}
		\rule{0pt}{10pt}
		 & Kunming & 56 & 8465\\
		\bottomrule
	
	\end{tabular}
	\caption{Statistics of high-risk neighborhoods of datasets.}
	\label{table-apdx-dataset}
\end{table}

\subsection{Hyperparameters}\label{apx:hyp}
We evaluate cross-city early detection performance of our C-Watcher and baselines in section \ref{sec: hyperparameters}. 
Table \ref{table-apdx-hyper-baseline} presents the hyperparameters of each baseline.

\subsection{Results of Early Detection}\label{apx:red}
Table \ref{table-apdx-results} shows the high-risk neighborhoods detecting performance our C-Watcher and baselines. The ``target city - reference city'' relationship are ``Huizhou \& Guangzhou -- Shenzhen'', ``Shaoyang \& Yiyang -- Changsha'', ``Lianyungang \& Nanjing -- Shanghai'', ``Xuchang \& Anyang -- Zhengzhou'' and `` Chongqing \& Kunming -- Chengdu''.

\subsection{Features Abbreviation and Full Name} \label{apx:fafn}
In the analysis of features importance in section \ref{features-importances}, we use abbreviation to name our constructed features (see section \ref{sec:feature}) for simplicity. The abbreviation -- full name matches are presented in Table \ref{table-apdx-fullname}.

\begin{table*}[!t]
    \centering
	\begin{tabular}{c|c}
		\toprule
		Types of Living Facilities & Types of POIs \\
		\midrule
		Medical Institutions & Comprehensive Hospital $|$ Special Hospital $|$ Clinic\\
		\hline
		\rule{0pt}{10pt}
		Shopping Places & Shopping mall $|$ Supermarket $|$ Market $|$ Shop\\
		\hline
		\rule{0pt}{10pt}
		Sports Venues & Gymnasium $|$ Fitting Center $|$ Extreme sports venues $|$ Others\\
		\hline
		\rule{0pt}{10pt}
		Education Services & Primary School \& Middle School \\
		\hline
		\rule{0pt}{10pt}
		Catering Services & Chinese Restaurant $|$ Western Restaurant \\
		\hline
		\rule{0pt}{10pt}
		Financial Services & Bank \\
		\hline
		\rule{0pt}{10pt}
		Communication services & Communication Business Hall\\
		\hline
		\rule{0pt}{10pt}
		Government Apparatus & Public Security Organ \\
		\hline
		\rule{0pt}{10pt}
		Transportation Facilities & Bus Stop $|$ Subway Station\\
		\bottomrule
		
\end{tabular}
\caption{Types of basic living facilities and corresponding types of POIs. ``\&'' represents that a neighborhood equips such type of living facility when all the types of POIs connected by ``\&'' should be within 1$km$, while ``$|$'' represents that a neighborhood equips such type of living facility as long as one of the types of POIs connected by ``$|$'' is within 1$km$.}
\label{table-apdx-facilities}
\end{table*}
\begin{table*}[!ht]
    \centering
	\begin{tabular}{c|c|c}
		\toprule
		Category & Feature & Value \\
		\midrule
		&  & Walk \\
		\cline{3-3}
		\rule{0pt}{10pt}
		\multirow{2}{*}{T} & \multirow{2}{*}{Means of Transportation} &  Bicycle \\
		\cline{3-3}
		\rule{0pt}{10pt}
		&  &  Public Transmit \\
		\cline{3-3}
		\rule{0pt}{10pt}
		&  &  Private Vehicle \\
		
		\hline
		\rule{0pt}{10pt}
		\multirow{6}[4]{*}{OD} &   & Hotel \\
		\cline{3-3}
		\rule{0pt}{10pt}
		& Types of Visited Venues &  Shopping \\
		\cline{3-3}
		\rule{0pt}{10pt}
		&  &  ... \\
		
		\cline{2-3}
		\rule{0pt}{10pt}
		 &   & $< 1km$ \\
		\cline{3-3}
		\rule{0pt}{10pt}
		& Origin--Destination Distance &  $2-5km$ \\
		\cline{3-3}
		\rule{0pt}{10pt}
		&  &  ... \\
		\hline
		\rule{0pt}{10pt}
		 & 	& Neighborhood -- Neighborhood \\
		\cline{3-3}
		\rule{0pt}{10pt}
		& Top-20 hottest Origin -- Walk -- Destination Patterns  &  Neighborhood -- Food \\
		\cline{3-3}
		\rule{0pt}{10pt}
		&  &  ... \\
		\cline{2-3}
		\rule{0pt}{10pt}
		 &	& Neighborhood -- Neighborhood \\
		\cline{3-3}
		\rule{0pt}{10pt}
		& Top-20 hottest Origin -- Bicycle -- Destination Patterns &  Neighborhood -- Shopping \\
		\cline{3-3}
		\rule{0pt}{10pt}
		\multirow{2}[1]{*}{OTD} &  &  ... \\
		\cline{2-3}
		\rule{0pt}{10pt}
		&  & Neighborhood -- Neighborhood \\
		\cline{3-3}
		\rule{0pt}{10pt}
		& Top-20 hottest Origin -- Public Transmit -- Destination Patterns &  Neighborhood -- Shopping \\
		\cline{3-3}
		\rule{0pt}{10pt}
		&  &  ... \\
		\cline{2-3}
		\rule{0pt}{10pt}
		 & 	& Neighborhood -- Neighborhood \\
		\cline{3-3}
		\rule{0pt}{10pt}
		& Top-20 hottest Origin -- Private Vehicle -- Destination Patterns &  Neighborhood -- Company \\
		\cline{3-3}
		\rule{0pt}{10pt}
		&  &  ... \\
		\bottomrule
		
\end{tabular}
\caption{Transportation-related features.}
\label{table-apdx-OTD}
\end{table*}
\begin{table*}[h]
    \centering
    \small
    \renewcommand\tabcolsep{5.0pt}
	\begin{tabular}{c|c|c|c|c|c|c|c|c|c|c}
		\toprule
		& \multicolumn{2}{c|}{Shenzhen} & \multicolumn{2}{c|}{Changsha} & \multicolumn{2}{c|}{Shanghai} & \multicolumn{2}{c|}{Zhengzhou} & \multicolumn{2}{c}{Chengdu} \\
		\cmidrule{2-11}
    	& Huizhou & Guangzhou & Shaoyang & Yiyang & Lianyunguang & Nanjing &  Xuchang & Anyang & Chongqing & Kunming\\
		\midrule
		SVM  & 0.7049 & 0.5510 & 0.5615 & 0.6039 & 0.6728 & 0.5623 & \textbf{0.7330} & 0.5170 & 0.5693 & 0.5235\\
		\hline
		\rule{0pt}{10pt}
		XGB & 0.6266 & 0.5392 & 0.5190 & \textbf{0.6744} & 0.6182 & \textbf{0.6033} & 0.7067 & 0.4604 & 0.4901 & 0.5717  \\
		\hline
		\rule{0pt}{10pt}
		Lasso-R & 0.6364 & 0.5798 & 0.5410 & 0.6318 & 0.6515 & 0.5556 & 0.7195 & 0.4394 & 0.5718 & 0.5263 \\
		\hline
		\rule{0pt}{10pt}
		MLP & 0.6995 & 0.5688 & 0.5594 & 0.6113 & 0.6850 & 0.5664 & 0.7278 & 0.4631 & 0.5438 & 0.5377 \\
		\hline
		\rule{0pt}{10pt}
		C-Watcher & \textbf{0.7352} & \textbf{0.6008} & \textbf{0.6433} & 0.6618 & \textbf{0.7218} & 0.5776 & 0.7312 & \textbf{0.5637} & \textbf{0.6142} & \textbf{0.6403} \\
		\bottomrule
	\end{tabular}
	\caption{Early detection performance comparison between C-Watcher and baselines on 10 cross-city datasets.}
	\label{table-apdx-results}
\end{table*}
\begin{table*}[!h]
    \centering
	\begin{tabular}{c|c|c|c}
		\toprule
		MLP & SVM & XGB & Lasso-R \\
		\midrule
		Dense Layer 1024 &  & Min Child Weight = 1 \\
		ReLU &  & Depth = 10\\
		Dropout = 0.5 & C = 1 & N-Tree = 160 \\
		Dense Layer 1 & Gamma = 0.0196 & Gamma = 0.025 & $\lambda = 3$ \\
		Sigmoid & Tolerance = 0.001 & Learning Rate = 0.0001 & Max Iteration = 3 \\
		\cline{1-1}
		\rule{0pt}{10pt}
		Optimizer = Adam & Kernel = RBF & Scale Positive Weight = 1.8 & Solver = LibLinear\\
		Learning Rate = 0.0001 & Max Iteration = No Limit & Sub-Sample = 0.7\\
		L2 Weight Decay = 0.001 &  & Colsample-Bytree = 0.7\\
		Batch Size = 16 &  &  Colsample-Bynode = 1\\
        \bottomrule
\end{tabular}
\caption{Types of baselines MLP, SVM, XGB and Lasso-R. In all the different baselines, random seeds are set to 2. We conduct baseline experiments with Python 3.8 except XGB using Python 2.7.}
	\label{table-apdx-hyper-baseline}
\end{table*}

\begin{table*}[!h]
    \centering
    \small
	\begin{tabular}{l|c|l|c}
		\toprule
	    Abbreviation & Full Name & Abbreviation & Full Name \\
	    \midrule
	    P : PFLF & Perfect Living Facilities & D : LFCL & Low Food Consumption Level \\
	    \hline
	    \rule{0pt}{10pt}
	    P : PRLF & Poor Living Facilities & D : LIL & Low Income Level \\
	    \hline
	    \rule{0pt}{10pt}
	    P : RTS  & Radius to Train Station & D : ACD & Average Commute Distance \\
	    \hline
	    \rule{0pt}{10pt}
	    \multirow{2}{*}{P : RPMSK} & Radius to Primary / Secondary School & \multirow{2}{*}{T : N-N-PV} & From Neighborhood to Neighborhood \\
	    & and Kindergarten &  & by Private Vehicle \\
	    \hline
	    \rule{0pt}{10pt}
	    P : RS & Radius to Shop & T : N-N-W & From Neighborhood to Neighborhood on Walk \\ 
	    \hline
	    \rule{0pt}{10pt}
	    P : RA & Radius to Airport & T : TW & Travel on Walk \\
	    \hline
	    \rule{0pt}{10pt}
	    P : RC & Radius to Campus & T : TD$>30$ & Travel Distance $>30km$\\
	    \hline
	    \rule{0pt}{10pt}
	    D : PD & Population Density & T : $2<$TD$<5$ & $2km<$ Travel Distance $<5km$\\
	    \hline
	    \rule{0pt}{10pt}
	    D : MFCL & Middle Food Consumption Level & T : N-T-B & From Neighborhood to Transportation by Bus\\
	    \hline
	    \rule{0pt}{10pt}
	    D : LEL & Low Educational Level & T : N-N-B & From Neighborhood to Neighborhood by Bus \\
	    \bottomrule
	\end{tabular}
	\caption{Abbreviation and corresponding full name of Features.}
	\label{table-apdx-fullname}
\end{table*}

\end{document}